\documentclass[10pt,twocolumn,letterpaper]{article}

\usepackage[final]{cvpr}      
\usepackage{multirow}
\usepackage{diagbox}
\usepackage{textcomp}

\usepackage{diagbox}
\usepackage{textcomp}
\usepackage[dvipsnames]{xcolor}

\definecolor{cvprblue}{rgb}{0.21,0.49,0.74}

\def\paperID{4} 
\def\confName{CVPM}
\def\confYear{2024}

\title{Toward Motion Robustness: A masked attention regularization framework in remote photoplethysmography}

\author{Pengfei Zhao$^{1}$ \thanks{corresponding author}\quad Qigong Sun$^{1,2,4}$\quad Xiaolin Tian$^{3}$\quad Yige Yang$^{3}$\\
Shuo Tao$^{3}$\quad Jie Cheng$^{3}$\quad Jiantong Chen$^{1}$\\
$^{1}$ Shanghai AI Laboratory \quad $^{2}$ SenseTime Research \quad $^{3}$ Xidian University \quad $^{4}$ Tsinghua University \\
{\tt\small $^{1}$zhaopengfei@pjlab.org.cn \quad $^{2}$sunqigong@sensetime.com \quad $^{1}$18591956639@163.com} \\
{\tt\small $^{3}$\{xltian@mail., yigeyang@stu., 23171214664@stu., 23171214606@stu.\}xidian.edu.cn}
}

\begin{document}
\maketitle
\thispagestyle{empty}

\begin{abstract}
    There has been growing interest in facial video-based remote photoplethysmography (rPPG) measurement recently, with a focus on assessing various vital signs such as heart rate and heart rate variability. Despite previous efforts on static datasets, their approaches have been hindered by inaccurate region of interest (ROI) localization and motion issues, and have shown limited generalization in real-world scenarios. To address these challenges, we propose a novel masked attention regularization (MAR-rPPG) framework that mitigates the impact of ROI localization and complex motion artifacts. Specifically, our approach first integrates a masked attention regularization mechanism into the rPPG field to capture the visual semantic consistency of facial clips, while it also employs a masking technique to prevent the model from overfitting on inaccurate ROIs and subsequently degrading its performance. Furthermore, we propose an enhanced rPPG expert aggregation (EREA) network as the backbone to obtain rPPG signals and attention maps simultaneously. Our EREA network is capable of discriminating divergent attentions from different facial areas and retaining the consistency of spatiotemporal attention maps. For motion robustness, a simple open source detector MediaPipe for data preprocessing is sufficient for our framework due to its superior capability of rPPG signal extraction and attention regularization. Exhaustive experiments on three benchmark datasets (UBFC-rPPG, PURE, and MMPD) substantiate the superiority of our proposed method, outperforming recent state-of-the-art works by a considerable margin.
\end{abstract}    
\section{Introduction}
\label{sec:intro}

\begin{figure}[t]
    \begin{center}
    \begin{tabular}{cc}
    \includegraphics[width=1.45in]{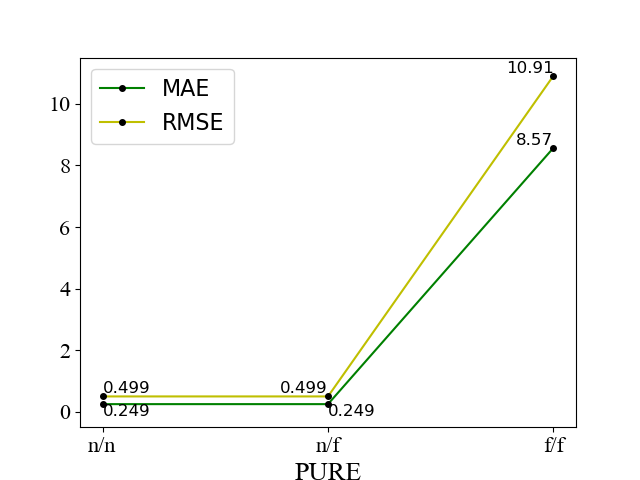} &
    \includegraphics[width=1.55in]{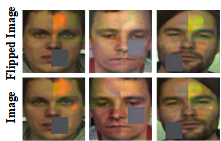} \\
    \scriptsize (a) HR evaluation (MAE) & \scriptsize (b) Attention comparison\\
    \end{tabular}
    \end{center}
    \vspace{-0.1in}
    \caption{The model degradation due to inaccurate and inconsistent ROI localizations on PURE dataset under the flip senmatic consistency strategy. `n' and `f' in (a) mean the normal data and horizontally flipped data, respectively. \textit{i.e.}, `n/f' denotes that the model trained with normal input and tested on flipped data. (b) shows the inconsistent attention map samples which are supposed to act as mirror attention regions.}
    \label{fig1}
  \end{figure}

Heart rate (HR) is a vital physiological signal that plays a significant role in various medical and health-related contexts. Its measurement is widely used to assess cardiovascular health, monitor physical activity, and diagnose diseases. Medical devices such as electrocardiography (ECG) and photoplethysmography (PPG)/Blood Volume Pulse (BVP) recorders can measure these signals through skin contact. However, these electrodes and wires need to be attached to body skins, probably causing discomfort and inconvenience to users, and even triggering allergic reactions, finally limiting their usefulness and scalability \cite{Alpher01, Alpher02}. In recent years, there has been a surge of interest in remotely measuring physiological signals from human facial videos captured by RGB cameras \cite{Alpher03, Alpher04, Alpher05}. This technology has various applications, such as healthcare in nursing homes \cite{Alpher06} and driver status assessment \cite{Alpher07}.

The majority of remote physiological measurement methods employ the principle of remote photoplethysmography (rPPG) \cite{Alpher13,Alpher14,Alpher15} , which tracks the periodic reflected ambient light changes in skin color caused by the variation in blood volume \cite{Alpher08,Alpher09,Alpher10,Alpher11,Alpher12}. Nonetheless, rPPG signals are susceptible and vulnerable to interference from non-periodic noise sources, including illumination changes of videos, facial expression changes, and body/head movements \cite{Alpher06,Alpher16,Alpher17,Alpher18,Alpher19}. Previous traditional studies have proposed blind signal separation techniques \cite{Alpher17,Alpher20,Alpher21} and skin reflection models \cite{Alpher08,Alpher22} to isolate rPPG signals from noises, consisting of multiple process such as rPPG extraction, filtering and post-processing. However, these approaches rely on certain assumptions, such as the assumption that the white light remains constant across individuals and no movement during video recording, which may not always hold true. That is to say, these methods are always tested on controlled ideal scenarios and show little robustness on illumination changes and head movements. As a result, these methods are expected to experience performance degradation in real-world scenarios.

Recently, there has been an increasing attention for applying deep learning techniques to remote photoplethysmography due to its potential to revolutionize health monitoring and diagnostics. Researchers have proposed various neural network models that leverage convolutional neural networks (CNNs) to extract spatiotemporal features from facial videos \cite{Alpher23,Alpher24,Alpher25,Alpher26,Alpher27}. These models have demonstrated their superior performance compared to traditional methods. However, there remains a critical challenge in accurately localizing the region of interest (ROI) of the face areas during the model training process \cite{Alpher28,Alpher29}. Noisy data, resulting from either different-view recordings and data transormations, or  illumination variations in real-world scenarios, can significantly affect the accuracy of ROI localization. This issue is particularly concerning since facial ROIs are known to change location in response to changes in blood volume, making it essential to identify their correct locations along with time for reliable rPPG measurements. Inaccurate ROI localization can result in the loss of valuable pulsation information, ultimately leading to poor generalization performance. Previous methods simply use learnable attention masks to assign weights to skin areas, like DeepPhys \cite{Alpher04}, but there is no guarantee for attention consistency in the view of spatial and temporal dimension. As illustrated in Fig.~\ref{fig1}, training only with horizontally flipped data suffers a lot from inaccurate ROI localizations, finally leading to model performance deterioration. To overcome this limitation, we introduce a novel masked attention regularization framework that enables accurate face ROI localization, thereby achieving reliable rPPG measurements.

Motion robustness is another challenge in remote photoplethysmography measurement, which is further exacerbated by the limitations of CNNs in handling spatial contextual ROI localization and motion-related issues. The relatively small receptive field of CNNs makes them susceptible to interference from various types of motions, including head movements and body language, which can significantly degrade the accuracy of rPPG measurement. While some existing end-to-end models have been tested on static datasets and neglected the motion robustness problem \cite{Alpher30}, they remain vulnerable when encountering realistic scenarios with noticeable head movements. Only a few research studies have attempted to improve motion robustness by incorporating ROI tracking or motion representation techniques, but these methods have shown mixed results, especially when dealing with more complex scenarios like walking or exercising \cite{Alpher31,Alpher32}. Therefore, developing effective motion robustness strategies remains an open challenge in rPPG research.

In this paper, we propose a novel masked attention regularization (MAR-rPPG) framework to guarantee attention consistency and alleviate motion impact for better rPPG measurement. Firstly, we introduce a masked attention regularization training framework to capture spatial and temporal attention consistency. Attention consistency means that the learned attention maps ought to conform to the identical transformation as the input images, thereby improving the model's ability to identify patterns and relationships between input features and labels. We find that flip semantic consistency is a simple but effective way, which means original video clips and their flipped counterparts should have similar attention maps. Furthermore, inspired by REA module in this work \cite{Alpher33}, we propose an enhanced rPPG expert aggregation (EREA) network to discriminately extract rPPG signals from different face regions and supervise the consistency of the original and flipped attention maps. Finally, unlike those complex and costly motion-robust solutions, we simply exploit the MediaPipe \cite{Alpher50} detector for preprocessing since it is enough for our excellent model to tackle with non-rigid/rigid motion issues. 

Overall, our MAR-rPPG has three key contributions:

\begin{enumerate}
   \item We introduce a masked attention regularization training framework that focuses on capturing both spatial and temporal attention consistency in remote photoplethysmography measurements. This framework enables the model to learn a robust pattern and representation from the input data. 
   \item We propose an enhanced rPPG expert aggregation (EREA) network paired with an attention consistency loss function to discriminately assign different attentions to face regions and supervise the consistency of the original and flipped attention maps.
   \item We conduct comprehensive experiments on three widely used benchmark datasets, namely UBFC-rPPG \cite{Alpher35}, PURE \cite{Alpher36}, MMPD \cite{Alpher37}. Experimental results demonstrate that our approach substantially outperforms the state-of-the-art methods under both simple and complex real-world scenarios.
\end{enumerate}

\section{Related works}
\label{sec:rwork}

Traditional facial video-based remote physiological measurement methods initially utilize blind signal separation techniques, such as independent component analysis (ICA) or principal components analysis (PCA) \cite{Alpher38,Alpher17,Alpher19}, and skin reflection models \cite{Alpher08,Alpher22,Alpher39,Alpher40}. Blind signal separation-based approaches assume that skin color changes are a linear combination of the desired rPPG signal and other noise sources. For instance, Macwan et al. \cite{Alpher39} utilized autocorrelation to guide ICA for separating rPPG signals, while McDuff et al. \cite{Alpher40} employed ICA to estimate rPPG signals from multiple facial videos captured at diverse angles. On the other hand, skin reflection model-based approaches aim to transform images from the RGB space to alternative color spaces for enhanced rPPG estimation. For example, Haan et al. \cite{Alpher08} projected images into the chrominance space to minimize motion noise and facilitate rPPG estimation. However, these methods rely on questionable assumptions, such as the notion that different individuals' skin tones appear identical under white light, which may not hold true in practical environments \cite{Alpher25,Alpher41}. A comprehensive review of these conventional techniques can be found in \cite{Alpher42}.

Recent studies have demonstrated that deep neural network (DNN)-based approaches can achieve superior performance in remote physiological measurement \cite{Alpher32,Alpher41,Alpher44,Alpher45}. Researchers have developed various DNN-based methods to address different challenges in remote physiological measurement. Notably, DeepPhys \cite{Alpher04} and PhysNet \cite{Alpher30} introduced end-to-end convolutional neural network (CNN) frameworks that have achieved promising results. Additionally, spatio-temporal signal map-based methods, such as those presented by RhythmNet \cite{Alpher43} and Dual-GAN \cite{Alpher46}, have gained popularity due to their impressive performance. In pursuit of efficient rPPG measurement, researchers have explored lightweight end-to-end models, such as Auto-HR \cite{Alpher47} and EfficientPhys \cite{Alpher31}. These models aim to reduce computational complexity while maintaining accuracy. Moreover, PhysFormer \cite{Alpher27} employs temporal difference transformers to investigate long-range spatio-temporal relationships in rPPG representations, further improving performance. Despite their high performance on rPPG measurements, these methods encounter difficulties in accurately localizing regions of interest (ROIs) and maintaining consistent attention throughout the time sequence of video clips.

To facilitate motion-robust research, Paruchuri et al. \cite{Alpher48} explored a data augmentation by transferring motion to existing datasets and then trained on these motion-augmented datasets to improve the model generalization. Li et al. \cite{Alpher49} suggested two modular components: the physiological signal feature extraction block (PFE) and the temporal face alignment block (TFA). These modules aim to mitigate the impact of varying distances and head movements on remote physiological measurement, thereby enabling motion-robust studies. However, integrating these modules significantly increases the computational load and time cost, posing considerable challenges for independent researchers or those with limited resources in the field. Instead, we only utilize the MediaPipe toolkit for face tracking preprocessing since our proposed method has the outstanding capability to extract rPPG signals from semi-processed data.

\section{Methodology}
\label{sec:method}

In this section, we introduce the details of our proposed solution toward motion robustness -- the Masked Attention Regularization (MAR-rPPG) framework.

\begin{figure*}[t]
  \centering
   \includegraphics[width=1.0\linewidth]{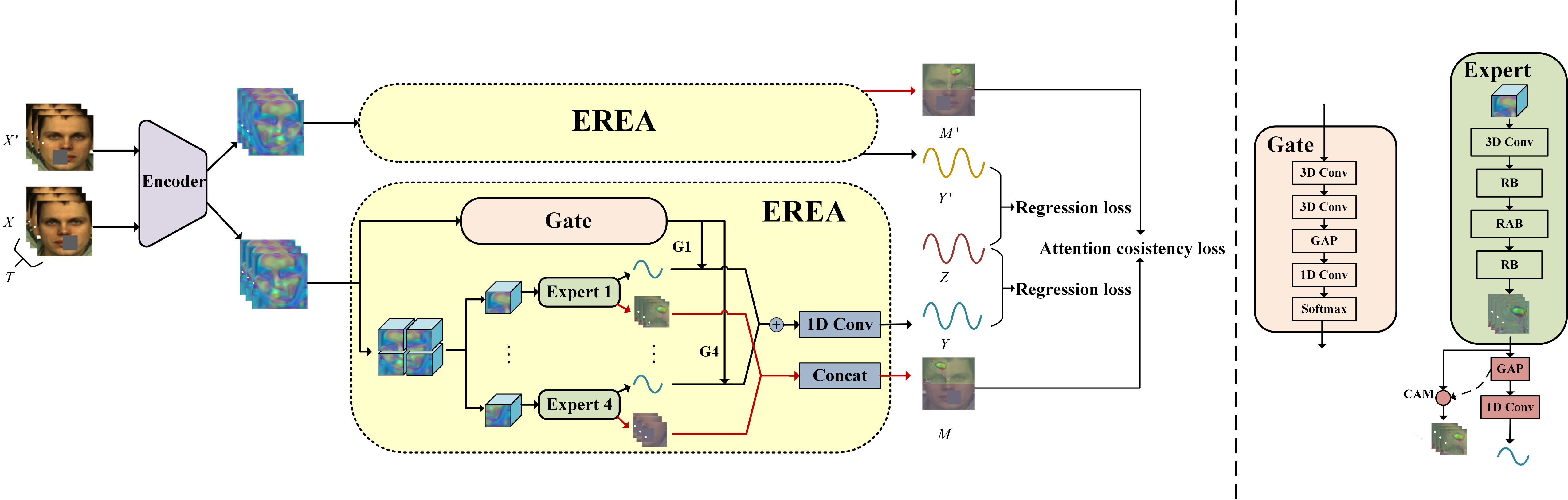}

   \caption{The overview of our proposed MAR-rPPG. The MAR-rPPG consists of one Encoder and one EREA network with shared weights for two inputs. First, the encoder encodes the input video into a feature tensor. Next, this feature tensor is sent to the EREA network and EREA outputs a rPPG signal and attention maps. Specifically in the EREA, the feature tensor is divided into four equal parts, and each part is pass to its corresponding Expert $E$ module to generate attention maps and extract a rPPG signal corresponding to one of facial regions. Finally, a Gate module $G$ aggregates four different rPPG signals into one rPPG prediction.}
   \label{fig2}
\end{figure*}

\subsection{Masked attention regularization}

In this paper, we propose a novel framework, called MAR-rPPG, designed to enhance rPPG signals by improved ROI localization. The overview of our proposed MAR-rPPG framework is illustrated in Fig.~\ref{fig2}, the proposed method consists of three main stages: feature extraction, signal and attention map generation, and network optimization. First, in the preprocessing process, we utilize the open-source face detector MediaPipe to detect, track, align, and crop the extended face area in each frame of the input video. Next, we create both original and flipped versions of the aligned facial frames, divided into chunks of $T$ frames each. Now we have original input $X$ and flipped input $X'$ and corresponding PPG signal label $Z$, we then pass these two input $X$ and $X'$ into the EREA backbone to generate corresponding rPPG signals and attention maps, denoted as $Y$, $M$, $Y'$, $M'$, respectively. Then, our EREA network (Details in Sec. 3.2) can obtain a rPPG prediction by aggregating pulsation information from diverse face regions and employing an attention regularization to guarantee accurate and consistent ROI locations. In the network optimization phase, we apply a regression loss to the signals from $Y$ and $Y'$, comprising $L1$ loss and negative Pearson loss. Additionally, we enforce attention consistency loss between attention maps from $M$ and $M'$, ensuring that identical ROI attentions are assigned to original and flipped inputs.

The underlying rationale for MAR-rPPG is grounded in the discernment that estimated heart rate values manifest dissimilarities pre- and post-flipping transformations, indicative of divergent ROI localizations during model training. Typically, rPPG signal extraction is confined to specific attention areas, posing a challenge for consistent localization. To overcome this, we integrate an attention regularization mechanism into the framework, ensuring attention homogeneity under prescribed image augmentations to enhance ROI localization within video sequences. Our method employs an indirect strategy to guide the backbone's attention towards regions exhibiting heightened regression correlation. 
Our MAR-rPPG introduces a global and dynamic attention regularization framework, which differs from the conventional practice that explicitly focuses on part of facial regions relevant to regression target during training, offering an indirect modality to improve network attention capability.

The attention regularization technique allows the model to produce consistent attention maps and rPPG signals, but it may not completely eliminate the negative effects of the flipped frames, as the model can still memorize the flipped version, leading to model performance degradation. To address this issue, we introduce a masking strategy. Before passing facial frames into model, we randomly mask part of facial area to prevent the model from overfitting on the flipped frames and encourage it to focus on the underlying patterns of input data.

\subsection{Enhanced rPPG expert aggregation network}

As illustrated in Fig.~\ref{fig2}, the enhanced rPPG expert aggregation (EREA) network is composed of four \textit{Expert} modules and one \textit{Gate} module. The feature tensor from \textit{Encoder} is divided into four equal parts, and each part is sent to its corresponding \textit{Expert} to extract the rPPG signal and attention maps. Then, the \textit{Gate} module aggregates four different rPPG signals from \textit{Experts} into one rPPG prediction.

The motivation of the EREA network is that different facial regions have distinct distributions of blood vessels and noises, and are supposed to contribute differently to the rPPG estimation. The EREA network serves as the backbone of MAR-rPPG framework and plays two roles in this pipeline: an attention map generator and a rPPG signal extractor. 

For one thing, we equip the vanilla \textit{Expert} module in \cite{Alpher33} with a Class Activation Mapping (CAM) unit to ensure the consistency of high-dimensional attention maps and visualizing the ROI locations on facial areas, since attention consistency assumes that learned attention maps are supposed to follow the identical transformation as input images, which benefits to achieve better visual explainability. Therefore, our EREA network is able to highlight discriminate ROIs relevant to pulsation information and enables the network to better understand the spatial relationships between different parts of face and its underlying physiological signals.

Specifically, an attention map is created by taking the weighted sum of the feature maps produced by the final convolutional layer, where the weights are derived from a fully connected (FC) layer. We denote the feature map as $ F\in \mathbb{R}^{C\times H\times W} $, where $C$, $H$, and $W$ represent the number of channels, height, and width of the feature map, respectively. Let $W\in \mathbb{R}^{T\times C}$ as the weights of the FC layer, $T$ is the regression dimension of rPPG signal chunks. The attention map is as below.

\begin{equation}
  M(h,w)=\sum_{c=1}^{C} W(T,c)F_{c}(h,w)
  \label{eq:attention_map}
\end{equation}
  where $c \in C$, $h \in H$, and $w \in W$.

For another, the generated attention maps are also beneficial for the rPPG signal extractor. They help the extractor learn more robust patterns and representations, which in turn enhances its ability to accurately detect and analyze the subtle changes in facial color that indicate cardiovascular activity. This is particularly useful in situations where the subject is experiencing complex motions or facial expressions, as the attention maps help the network to selectively focus on the most relevant regions of the face and ignore irrelevant information.

\subsection{Network optimization}

In this section, we present two key losses of our network optimization strategy: regression loss and attention consistency loss. These two losses play a crucial role in enhancing the accuracy and robustness of our rPPG model.

\subsubsection{Regression loss}

Our goal is to recover rPPG signals with similar trends and pulse peak timings that coincide with the ground truth signals. To achieve this, we employ $L1$ loss $loss_{l1}$ and negative Pearson correlation loss $loss_{\rho}$ as the regression loss $loss_{reg}$ to optimize the trend similarity and minimize peak location errors, thereby ensuring the recovered rPPG signals closely mirror the ground truth signals.

\begin{equation}
  loss_{l1} = \frac{1}{T}\sum_{i=1}^{T}\left| y_{i}-z_{i} \right|
  \label{eq:l1_loss}
\end{equation}

\begin{equation}
  \small loss_{\rho} = 1- \frac{T\sum_{}^{}yz-\sum_{}^{}y\sum_{}^{}z}{\sqrt{(T\sum_{}^{}y^{2}-(\sum_{}^{}y)^{2})(T\sum_{}^{}z^{2}-(\sum_{}^{}z)^{2})}}
  \label{eq:pearson_loss}
\end{equation}

\begin{equation}
  loss_{reg} = (1-\alpha) loss_{l1} + \alpha loss_{\rho}
  \label{eq:regression_loss}
\end{equation}

where $T$ is the chunk length of rPPG signals, $y\in Y, Y'$ is the estimated rPPG signals, $z\in Z$ is the ground truth PPG signals, and $\alpha$ is the weight to tradeoff these two losses. 

\subsubsection{Attention consistency loss}

In addition to the regression loss, we also incorporate an attention consistency loss $loss_{ac}$ into our optimization process to ensure that the model's attention mechanism produces consistent results across frames of original and its flipped versions, further improving the quality and reliability of the recovered rPPG signals. The attention consistency loss is computed as

\begin{equation}
  loss_{ac} = \frac{1}{TCHW}\sum_{i=1}^{T} \sum_{j=1}^{C}\left\| M_{ij}-M'_{ij} \right\|_{2}
  \label{eq:attention_loss}
\end{equation}

Overall, the total loss is computed as follows

\begin{equation}
  loss = (1-\beta)loss_{reg} + \beta loss_{ac}
  \label{eq:attention_loss}
\end{equation}

where $\beta$ is the tradeoff weight for rPPG signals and attention maps.

\section{Experiments}
\label{sec:experiment}

\subsection{Datasets}

We conduct our experiments on three public datasets: PURE, UBFC-rPPG, MMPD.

\textbf {PURE}. The PURE dataset comprises 60 facial videos of 10 participants, captured during a data collection process that involved asking the subjects to perform six types of head movements (including small and medium rotations, slow and fast translations, talking, and steady positions) in front of a camera for one minute. The videos were recorded at a frame rate of 30 frames per second and a resolution of 640 x 480 pixels. Simultaneously, the ground truth PPG signals were collected using a finger clip pulse oximeter with a sampling rate of 60 Hz.

\textbf {UBFC-rPPG}. The UBFC-rPPG dataset includes 42 high-quality facial videos, each featuring a simultaneous recording of both PPG signals and heart rates. Each video boasts a crisp resolution of 640x480 pixels and a smooth frame rate of 30 frames per second (FPS).

\textbf {MMPD}. Introducing the Multi-domain Mobile Video Physiology Dataset (MMPD), an extensive repository of 11 hours of Samsung Galaxy S22 Ultra mobile phone recordings from 33 participants. Captured at 30 frames per second and originally rendered at $1280 \times 720$ pixels, the videos were later scaled down to $320 \times 240$ pixels for efficient storage and transfer. In parallel, the accompanying PPG signals were downsampled from 200Hz to 30Hz, matching the video frame rate and yielding 1800 frames per video.  It contains Fitzpatrick skin types 3-6, four different lighting conditions (LED-low, LED-high, incandescent, natural), four various activities (stationary, head rotation, talking, and walking), and exercise scenarios. To foster exploration of various factors influencing remote photoplethysmography, MMPD meticulously assigned diverse labels, encompassing skin tone, gender, glasses, hair coverage, and makeup.

\subsection{Implementation Details}

 For each video, we use the MediaPipe \cite{Alpher50} toolkit to detect and crop the enlarged face area and then rescale them to a resolution of 64x64 pixels. We then divide each video clip into chunks of 60 frame length. We also flip images in each chunk as flipped input for attention consistency learning. It's worth noting that the MMPD dataset scales the videos to a resolution of $320 \times 240$, which can result in a distorted image with an incorrect aspect ratio. As such, it's necessary to resize the frames from MMPD to $320 \times 180$ before using them for training or evaluation purposes. To the best of our knowledge, we are the first to successfully train a model on the MMPD dataset and achieve satisfactory results. 
 
 Furthermore, we train our model for 30 epochs with two NVIDIA GeForce RTX 4090 GPUs using PyTorch 1.12.1, employing the Adam optimizer and setting the batch size to 4. Our initial learning rate is 1e-3 and the OneCycle scheduler mechanism is utilized in our experiments. we set the random mask size to $16 \times 16$ pixels, and hyperparameter $\alpha$ is set to 0.3 and $\beta$ is 0.5. Besides, the chunk length $T$ is set to 60. We implement our proposed MAR-rPPG framework based on open source rppg-toolbox \cite{Alpher34}.

 \subsection{Metrics and evaluation}

We follow the approach outlined in \cite{Alpher34} by using mean absolute error (MAE), root mean square error (RMSE), mean absolute percentage error (MAPE), and Pearson’s correlation coefficient (r) as evaluation metrics for heart rate (HR). Moreover, we perform cross-dataset HR evaluation among PURE, UBFC-rPPG, and MMPD. For details, refer to Sec. 1 of the Supplementary Material.






 \subsection{Results}
 \subsubsection{HR evaluation}

For HR evaluation, we include eight advanced approaches: Meta-rPPG \cite{Alpher44}, PulseGan \cite{Alpher41}, Dual-GAN  \cite{Alpher46}, Physformer \cite{Alpher27}, REA-LFA \cite{Alpher33}, EfficientPhys \cite{Alpher31}, 
TS-CAN \cite{Alpher32}, PhysNet \cite{Alpher30}. We validate their model performance on three datasets as shown in Tab.~\ref{tab1}. It is worth noting that we did not include certain methods in our experiments on the MMPD(s) dataset, as their reproduction was not feasible due to various reasons such as unavailable code or inconsistent implementation details.

We observe that our proposed MAR-rPPG method outperforms all compared methods across all evaluated metrics. Notably, it achieves a MAE of 0.08 and a RSME of 0.29, vastly superior than the second-best performer Dual-GAN on the PURE dataset, which has a MAE of 0.82 and a RMSE of 1.31. Additionally, our method obtains a perfect Pearson's correlation coefficient ($r$) of 1.00 on both the UBFC-rPPG and PURE datasets, signifying an exact match between the predicted and actual heart rates. Of particular note is the impressive performance on the challenging MMPD dataset captured using a mobile phone, which includes diverse scenarios such as glasses, makeup, and varying skin tones. Despite these complications, our method maintains a remarkably low MAE of 0.87 on the MMPD(s) dataset, showcasing the efficacy of our masked attention regularization framework in enhancing ROI localization and bolstering the resilience of the rPPG model.

\begin{table*}[!t]
    \vspace{-0.1in}
    \caption{Comparison of state-of-the-art methods on HR estimation. The results are tested on UBFC-rPPG, PURE,  and MMPD datasets. $\uparrow$ indicates that the larger value is better and $\downarrow$ vice versa. The best metric is marked in \textbf{bold}. MMPD(s) means that we only test on the stationary scenario of this dataset for fair comparisons with other methods.}
    \begin{center}
    \begin{tabular}{c|ccc|ccc|ccc}
    \toprule
    \multirow{2}*{Method} &\multicolumn{3}{c|}{UBFC-rPPG}& \multicolumn{3}{c|}{PURE}& \multicolumn{3}{c|}{MMPD(s)}\\ 
    &MAE$ \downarrow$  &RMSE$ \downarrow$  &$r \uparrow$  & MAE$ \downarrow$  &RMSE$ \downarrow$  &$r \uparrow$ & MAE$ \downarrow$  &RMSE$ \downarrow$  &$r \uparrow$ \\ 
    \midrule
    Meta-rPPG \cite{Alpher44}   &5.97  &7.42  &0.53  &2.52  &4.63 &0.98  &6.05  &9.89  &0.67\\
    PulseGan \cite{Alpher41} &1.19 &2.10  &0.98  &2.28  &4.29 &0.99  &- &-  &- \\ 
    Dual-GAN  \cite{Alpher46}  &0.44  &0.67  &0.99  &0.82 &1.31  &0.99 &-  &-  &- \\
    Physformer \cite{Alpher27} &0.40 &0.71  &0.99  &1.10  &1.75 &0.99  &4.94 &9.31  &0.64\\ 
    REA-LFA \cite{Alpher33}  & 0.58 & 0.94 & 0.99 & 1.23 & 2.01 & 0.99 & - & - & -\\
    EfficientPhys \cite{Alpher31} & 1.14 & 1.81 & 0.99 & 1.33 & 5.99 & 0.97 & 1.32 & 2.67 & 0.97\\
    TS-CAN \cite{Alpher32} & 1.70 & 2.72 & 0.99 & 2.33 & 3.71 & 0.98 & 0.93 & 2.04 & 0.98\\
    PhysNet \cite{Alpher30} & 2.95 & 3.67 & 0.97 & 2.16 & 2.70 & 0.99 & 5.30 & 11.86 & 0.35\\
    \midrule
    MAR-rPPG(Ours) & \textbf{0.12} & \textbf{0.35} & \textbf{1.00} & \textbf{0.08} & \textbf{0.29} & \textbf{1.00} & \textbf{0.87} & \textbf{1.41} & \textbf{0.99}\\

    \bottomrule
    \end{tabular}
    \end{center}
    \label{tab1}
    \end{table*}

\subsubsection{Cross dataset evaluation}

In order to evaluate the robustness and adaptability of our proposed method, we conducted a comprehensive cross-dataset examination on the UBFC-rPPG, PURE, and MMPD datasets. The results of this experiment are presented in Tab.~\ref{tab3}. We conduct four cross-dataset tests, and MMPD(s)$\rightarrow$UBFC-rPPG means the metric trained on MMPD(s) and tested on UBFC-rPPG. Our proposed MAR-rPPG outperforms other approaches in terms of overall performance and generalization ability. Specifically, while Physformer achieves impressive results on the UBFC-rPPG dataset, its performance deteriorates significantly when applied to the PURE dataset. In contrast, our method maintains consistent excellence across both datasets, indicating its robustness and adaptability to diverse scenarios. Additionally, Dual-GAN displays notable performance in certain aspects, such as RMSE and correlation coefficient ($r$). The cross-dataset evaluation conclusively demonstrates the exceptional generalization capability of MAR-rPPG, empowering it to excel in unanticipated scenarios.

\begin{table*}[!t]
    \vspace{-0.1in}
    \caption{Comparison of state-of-the-art methods on cross-dataset HR estimation. \textit{i.e.}, MMPD(s)$\rightarrow$UBFC-rPPG means the metric trained on MMPD(s) and tested on UBFC-rPPG.}
    \begin{center}
    \setlength{\tabcolsep}{1.5mm}
    \begin{tabular}{c|ccc|ccc|ccc|ccc}
    \toprule
    \multirow{2}*{Method} &\multicolumn{3}{c|}{MMPD(s)$\rightarrow$UBFC-rPPG} &\multicolumn{3}{c|}{MMPD(s)$\rightarrow$PURE}&\multicolumn{3}{c|}{PURE$\rightarrow$UBFC-rPPG}&\multicolumn{3}{c}{UBFC-rPPG$\rightarrow$PURE}\\ 
    &MAE$\downarrow$  &RMSE$\downarrow$  &$r\uparrow$  & MAE$\downarrow$  &RMSE$\downarrow$  &$r\uparrow$ & MAE$\downarrow$  &RMSE$\downarrow$  &$r\uparrow$ & MAE$\downarrow$  &RMSE$\downarrow$  &$r\uparrow$\\ 
    \midrule
    Meta-rppg \cite{Alpher44}  &13.08 &19.75 &0.51 &9.07  &19.80  &0.56 &6.11 &7.58 &0.66 &4.00 &5.98 &0.92\\
    PulseGan \cite{Alpher41} &- &- &- &-  &-  &-   &2.30 &3.50 &0.97 &3.36 &5.11 &0.95 \\
    Dual-Gan \cite{Alpher46} &- &- &- &-  &-  &-  &2.03 &\textbf{3.01} &\textbf{0.97} &1.81 &\textbf{2.97} &\textbf{0.99} \\
    Physformer \cite{Alpher27} &\textbf{1.46} &\textbf{2.99}  &\textbf{0.98} & 11.22 & 21.25 & 0.60 &1.93 &3.02 &0.97 &1.99 &3.28 &0.99 \\
    \midrule
    MAR-rPPG(Ours) &1.74 &6.34 &0.92 &\textbf{1.46}  &\textbf{5.13}  &\textbf{0.98} &\textbf{1.67} &6.31 &0.93 &\textbf{1.28} &7.69 &0.95\\ 
    
    \bottomrule
    \end{tabular}
    \end{center}

    \label{tab3}
\end{table*}

\begin{figure}[t]
    \begin{center}
    \begin{tabular}{cc}
    \includegraphics[width=1.6in]{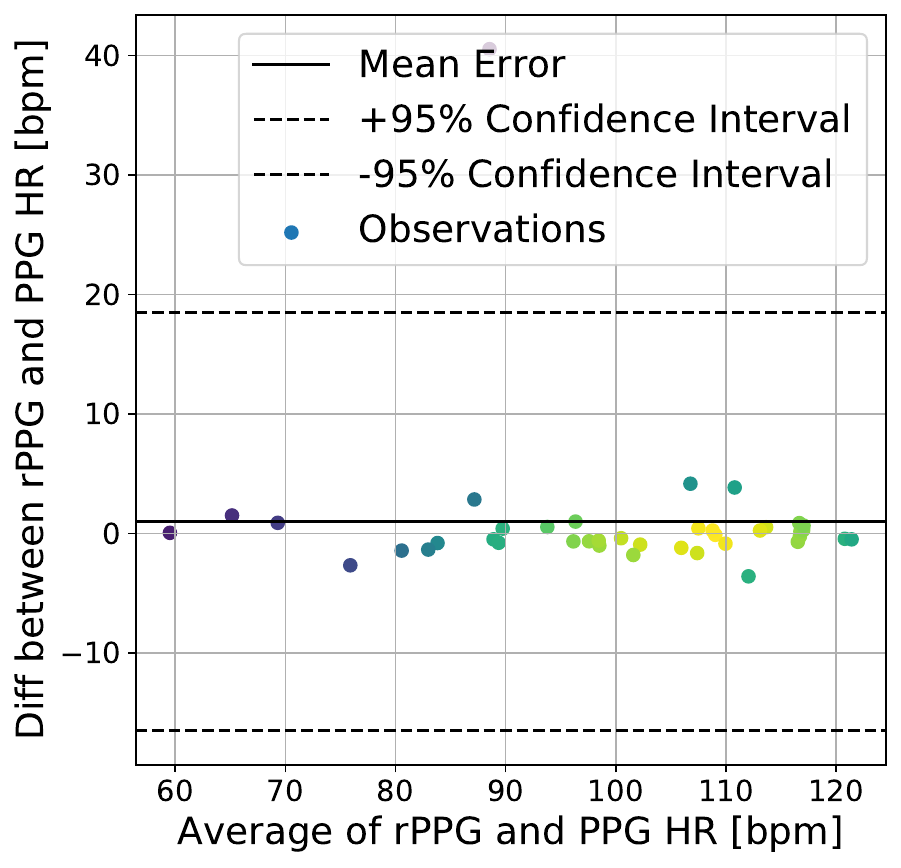} &

    \includegraphics[width=1.6in]{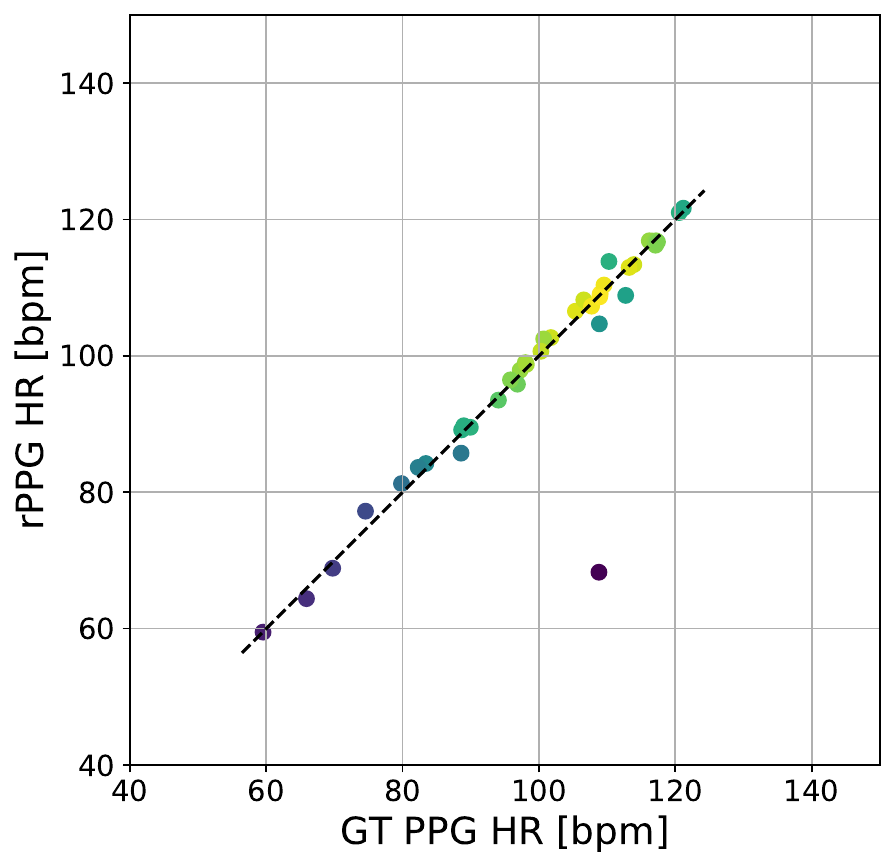} \\
    (a) & (b) \\
    
    \end{tabular}
    \end{center}
    \vspace{-0.1in}
    \caption{The Bland-Altman plot (a) and scatter plot (b) show the difference between estimated HR and ground truth HR on the crossdataset evaluation (PURE $\rightarrow$ UBFC-rPPG).}
    \label{fig3}
\end{figure}

\begin{figure}[t]
  \begin{center}
  \begin{tabular}{cc}
  \includegraphics[width=1.6in]{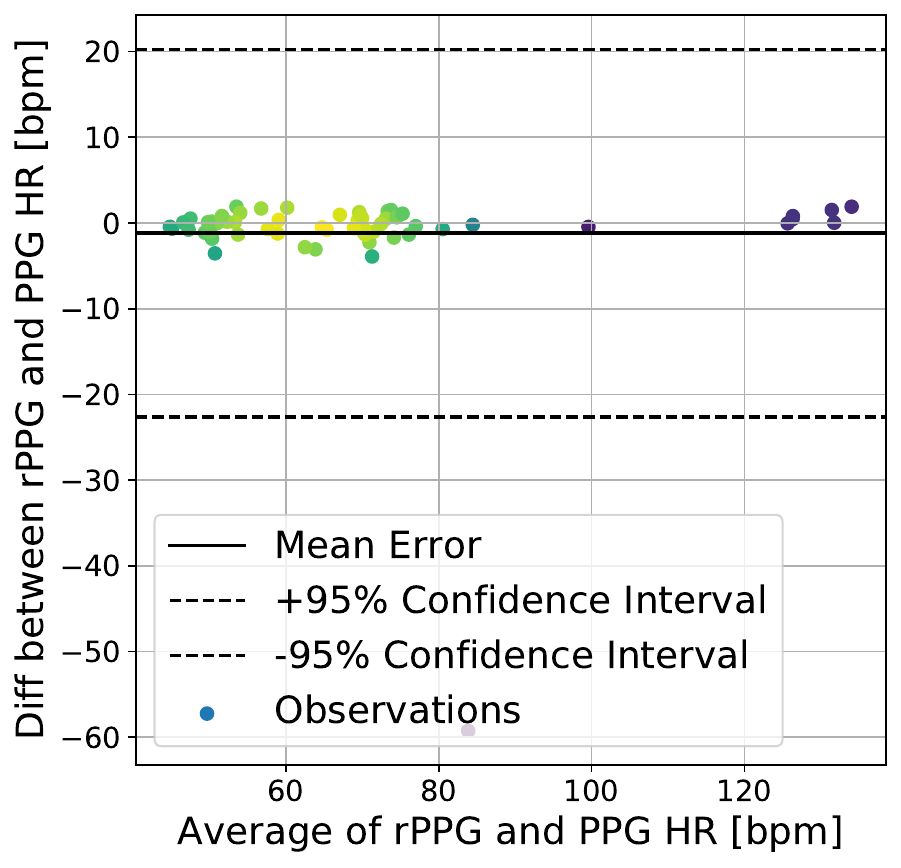} &

  \includegraphics[width=1.6in]{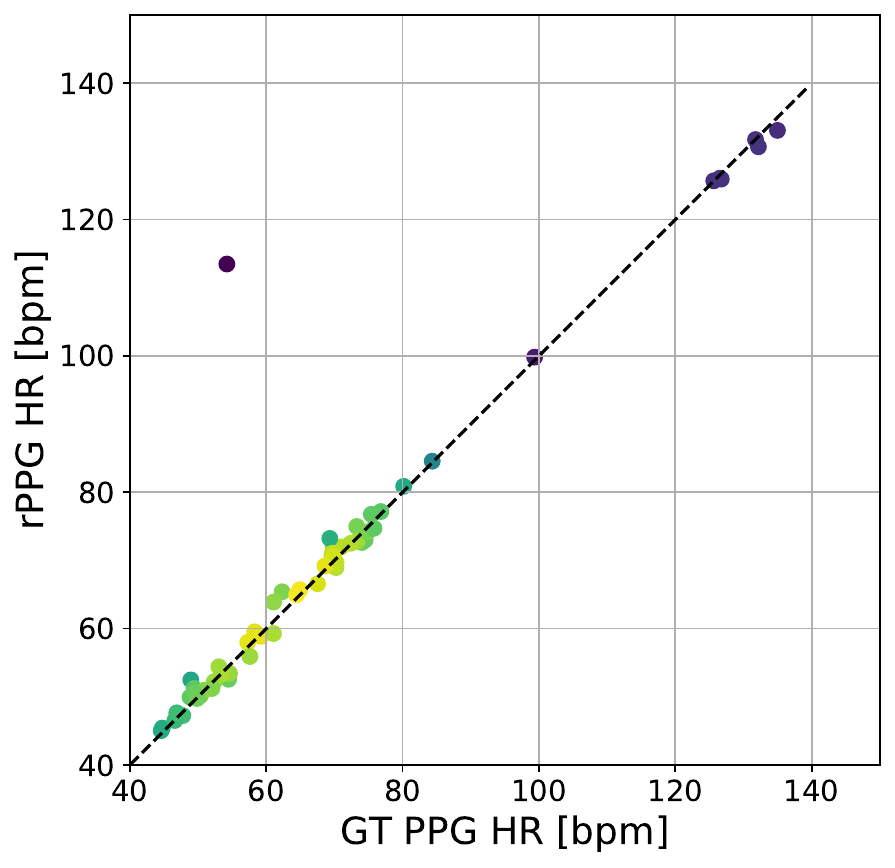} \\
  (a) & (b) \\
  
  \end{tabular}
  \end{center}
  \vspace{-0.1in}
  \caption{The Bland-Altman plot (a) and scatter plot (b) show the difference between estimated HR and ground truth HR on the crossdataset evaluation (UBFC-rPPG $\rightarrow$ PURE).}
  \label{fig4}
\end{figure}

Fig.~\ref{fig3} and ~\ref{fig4} display both a Bland-Altman plot and a scatter plot tested on UBFC-rPPG and PURE datasets. These plots allow for a comprehensive evaluation of the correlation between the ground truth PPG signals and the estimated rPPG signals. Each data point symbolizes an estimated result of a single test sample. For one thing, the Bland-Altman plot facilitates the identification of any potential systematic bias or random errors present in the estimation process. For another, the scatter plot enables a thorough examination of how well the estimated values align with the actual ones, thereby providing a better understanding of the model's precision. 

From Fig.~\ref{fig3} and ~\ref{fig4}, we can obviously notice that there is only one outlier for these two datasets while the rest points fit pretty well. So this is a foregone result that we surpass other state-of-the-art methods with a MAE of 1.67 under the test of PURE$\rightarrow$UBFC-rPPG and 1.28 for UBFC-rPPG $\rightarrow$ PURE.

\section{Ablation study}
\subsection{Attention regularization}

\begin{table*}[t]
  \centering
  \caption{Attention regularizaiton study on three dataset. MMPD(s) means that we remove the motion-related samples for a simple scenario, while MMPD(f) is nearly a full and complex dataset only without skin type 5 \& 6.}
  \begin{tabular}{ lc | lc lc lc lc  lc }
    \toprule
    Method & Dataset & MAE$\downarrow$ &  RMSE$\downarrow$ &  MAPE$\downarrow$ & $r\uparrow$ \\
    \midrule
    \multirow{4}{*} {vanilla REA \cite{Alpher33}} & UBFC-rPPG & 0.49 & 0.99 & 0.54 & 0.99 \\
    & PURE & 0.49 & 1.22 & 0.66 & 0.99\\
    & MMPD(s) & 0.99 & 1.57 & 1.22 & 0.99 \\
    & MMPD(f) & 4.81 & 12.53 & 4.86 & 0.60 \\
    \hline
    \multirow{4}{*} {MAR-rPPG(Ours)} & UBFC-rPPG & 0.12 & 0.35 & 0.11 & 1.00 \\
    & PURE & 0.08 & 0.29 & 0.11 & 1.00 \\
    & MMPD(s) & 0.87 &  1.65 & 1.02 & 0.99 \\
    & MMPD(f) & \textbf{2.83} & \textbf{8.18} & \textbf{2.94} & \textbf{0.82} \\
    \bottomrule
  \end{tabular}
  \label{tab4}
\end{table*}

Tab.~\ref{tab4} depicts the comparison of the vanilla REA and our proposed MAR-rPPG on three public datasets. The results demonstrate that MAR-rPPG outperforms vanilla REA across all metrics, a much lower MAE, RMSE, MAPE and a nearly perfect pearson correlation of 1.0 on UBFC-rPPG and PURE datasets. Notably, MAR-rPPG shows remarkable superiority on the more challenging MMPD(s) and MMPD(f) datasets, with a MAE of 0.87 in the static scenario, surpassing the 0.99 achieved by vanilla REA. Notably, while the vanilla REA struggles to handle the motion scenarios presented on the MMPD(f) dataset with a MAE of 4.81, our MAR-rPPG method demonstrates superior performance and maintain an acceptable MAE of 2.83. This suggests that our proposed MAR-rPPG is capable of effectively addressing the challenges posed by motion artifacts. 

\begin{figure}[t]
    \begin{center}
    \begin{tabular}{cc}
    \includegraphics[width=3.2in]{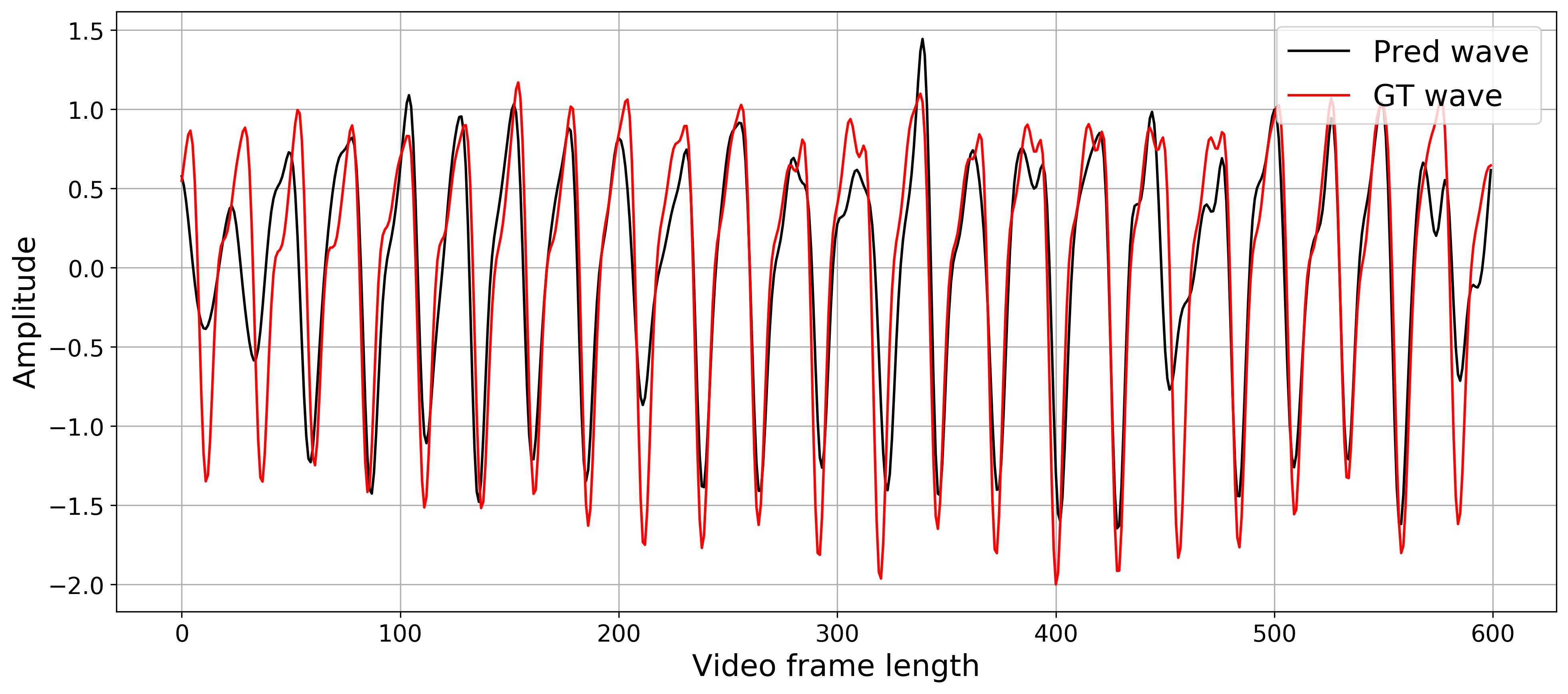} \\
    \scriptsize (a) Talking sample\\
    \includegraphics[width=3.2in]{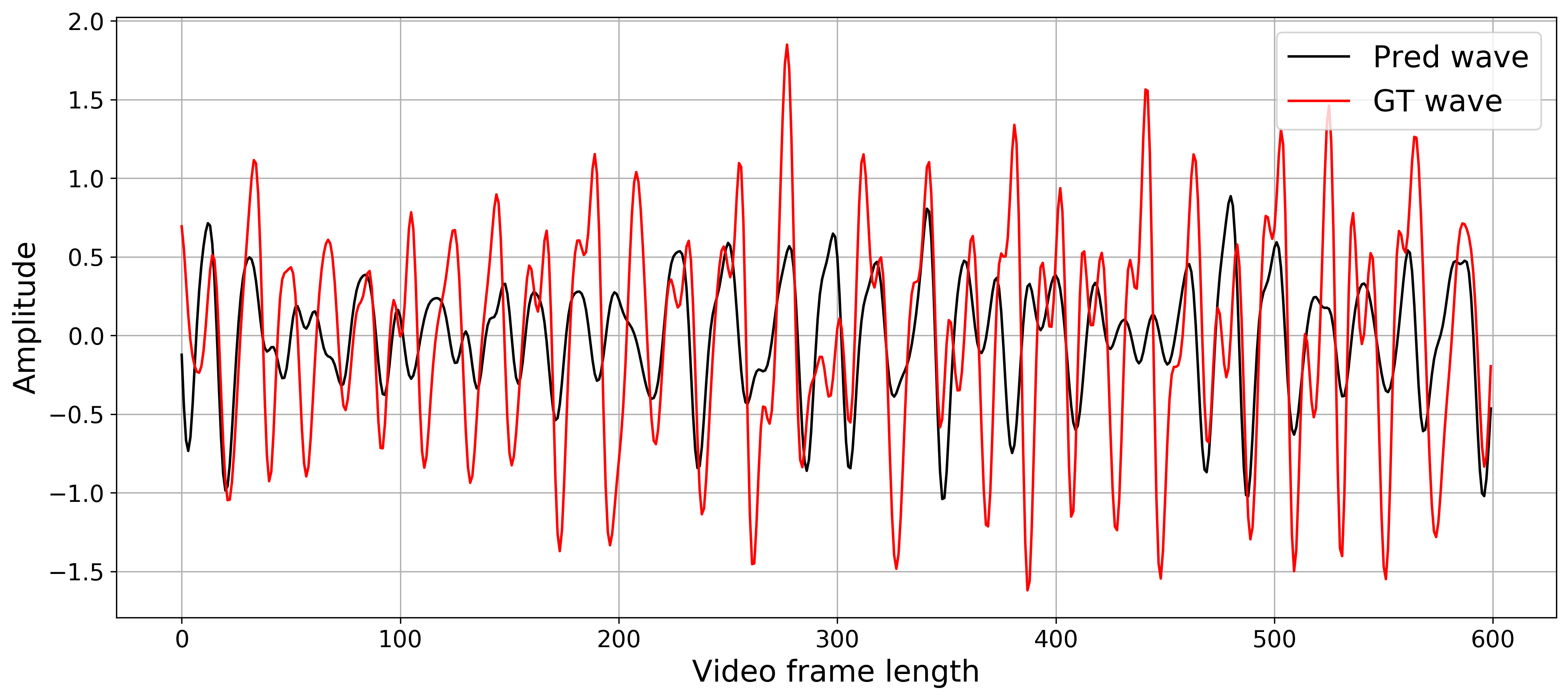} \\
    \scriptsize (b) Walking sample\\
    \end{tabular}
    \end{center}
    \vspace{-0.1in}
    \caption{The estimated rPPG signals on talking and walking motion samples of MMPD dataset. }
    \label{fig5}
  \end{figure}

As illustrated in Fig.~\ref{fig5}, we evaluate the motion-robust performance of our MAR-rPPG method on two scenarios: non-rigid motion (talking) and rigid motion (walking). Our method accurately estimates the rPPG signals for both scenarios, as demonstrated by the close match between the predicted and ground truth signals. In particular, the interbeat intervals of the predicted rPPG signals for the talking scenario are nearly identical to those of the ground truth PPG signal, indicating precise estimation. Moreover, our MAR-rPPG  successfully captures every waveform of the rPPG signal during the walking scenario and accurately predicts the heart rate.

  \subsection{The length of Video clips}

\begin{figure}[t]
  \begin{center}
  \begin{tabular}{cc}
  \includegraphics[width=1.65in]{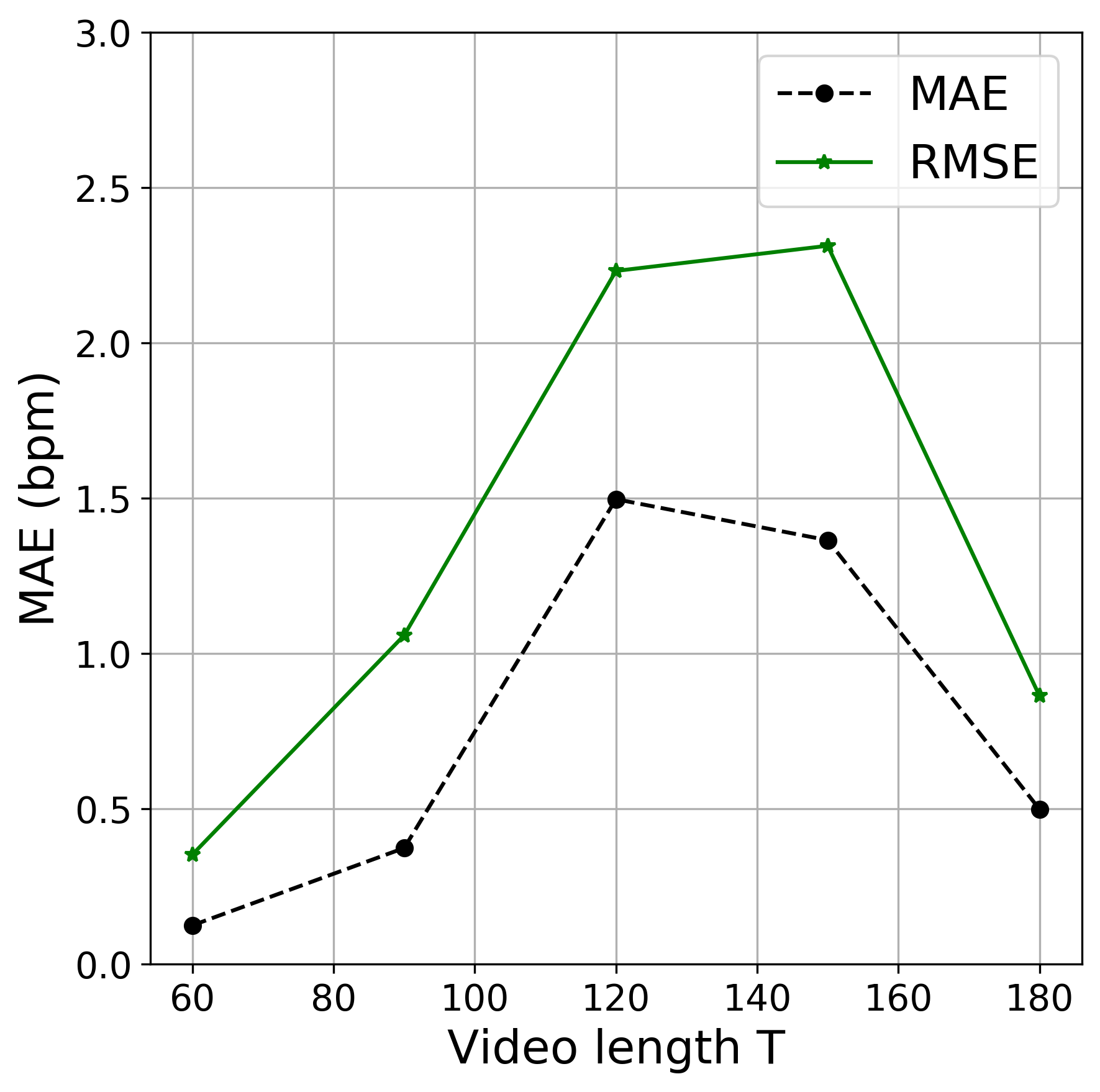} &
  \includegraphics[width=1.65in]{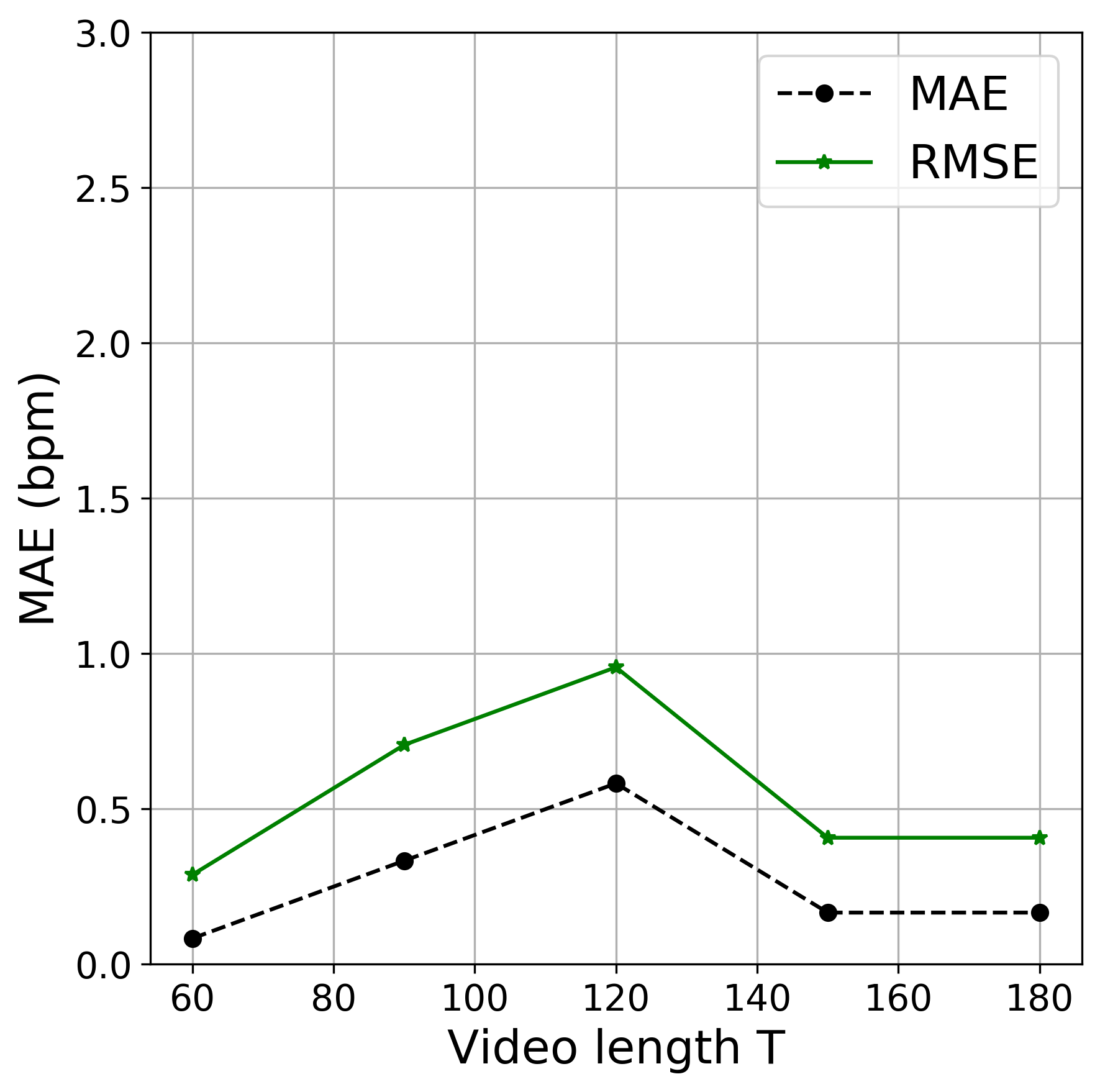} \\
  \scriptsize (a) UBFC-rPPG & \scriptsize (b) PURE\\
  \end{tabular}
  \end{center}
  \vspace{-0.1in}
  \caption{Parameter variation of chunk length. }
  \label{fig6}
\end{figure}

  We vary the length of each video clip $T$ from 60, 90, 120, 150, and 180 in Fig.~\ref{fig6}. We can observe that $T=60$ appears to be the best on UBFC-rPPG and PURE. The performance on PURE is rather stable along with varying $T$. One thing should be noted that larger chunk length means more GPU consumption, but with faster training speed if we fix the batch size.

\section{Conclusion}
In this paper, we propose a novel masked attention regularization framework for remote photoplethysmography measurement. The proposed method effectively alleviates the impact of inaccurate ROI localization and complex motion artifacts in non-contact rPPG measurements. First, we integrate a masked attention regularization mechanism to guarantee the visual semantic consistency of facial clips, which helps to reduce the impact of ROI localization errors. Additionally, we employ a masking technique to prevent the model from overfitting on inaccurate ROIs, thereby improving its generalization ability to unseen data. Second, we propose an Enhanced rPPG Expert Aggregation (EREA) network as the backbone to jointly learn rPPG signals and attention maps. The EREA network is designed to distinguish diverse attentions from different facial areas and preserve the consistency of spatiotemporal attention maps, which improves the motion robustness. Our extensive experiments demonstrate that MAR-rPPG outperforms state-of-the-art methods in terms of accuracy and motion robustness, making it a promising solution for real-world applications.

\section*{Acknowledgments}

This work was supported by the Shanghai Artificial Intelligence Laboratory. This research was conducted in collaboration with Xidian University - SenseTime Joint Lab of Smart Health Care.

{
    \small
    \bibliographystyle{ieeenat_fullname}
    \bibliography{main.bib}
}


\end{document}